\documentclass{article}
\usepackage{spconf,amsmath,graphicx}
\usepackage[mathscr]{eucal}
\usepackage{amssymb}
\usepackage{caption}
\usepackage{wrapfig}
\usepackage{marvosym}
\usepackage{graphicx}
\usepackage[backref]{hyperref}
\usepackage{multirow}
\usepackage{multicol}
\usepackage{floatrow}
\usepackage{etoolbox}
\usepackage{appendix}

\title{A comprehensive framework for occluded human pose estimation}
\name{Linhao Xu\textsuperscript{1}, Lin Zhao\textsuperscript{1(\Letter)}\thanks{(\Letter) Correspondence
should be sent to Lin Zhao (linzhao@njust.edu.cn)}, Xinxin Sun\textsuperscript{1}\thanks{Copyright © 2024 IEEE International Conference on Acoustics, Speech and Signal Processing (2024.ieeeicassp.org). All rights reserved.}, Di Wang\textsuperscript{2}, Guangyu Li\textsuperscript{1}, Kedong Yan\textsuperscript{1}}
\address{1. Nanjing University of Science and Technology, Nanjing, China\\
2. Xidian University, Xi'an, China}

\begin{document}
%
\maketitle
\begin{abstract}
Occlusion presents a significant challenge in human pose estimation. The challenges posed by occlusion can be attributed to the following factors: 1) Data: The collection and annotation of occluded human pose samples are relatively challenging. 2) Feature: Occlusion can cause feature confusion due to the high similarity between the target person and interfering individuals. 3) Inference: Robust inference becomes challenging due to the loss of complete body structural information. The existing methods designed for occluded human pose estimation usually focus on addressing only one of these factors. In this paper, we propose a comprehensive framework DAG (Data, Attention, Graph) to address the performance degradation caused by occlusion. Specifically, we introduce the mask joints with instance paste data augmentation technique to simulate occlusion scenarios. Additionally, an Adaptive Discriminative Attention Module (ADAM) is proposed to effectively enhance the features of target individuals. Furthermore, we present the Feature-Guided Multi-Hop GCN (FGMP-GCN) to fully explore the prior knowledge of body structure and improve pose estimation results. Through extensive experiments conducted on three benchmark datasets for occluded human pose estimation, we demonstrate that the proposed method outperforms existing methods. Code and data will be publicly available.

\end{abstract}
\vspace{-5pt}
\section{Introduction}
\vspace{-5pt}
\label{sec:intro}
Human pose estimation (HPE) has been a prominent area of research in computer vision, with the primary goal of accurately localizing annotated keypoints of human body, such as wrists and eyes, etc. This fundamental task serves as a basis for numerous downstream applications, including human action recognition\cite{ni2017learning},human-computer interaction \cite{mazhar2018towards} and pedestrian re-identification\cite{li2021diverse}, etc.

\begin{figure}[htbp!]
  \centering
  \includegraphics[width=0.9\linewidth]{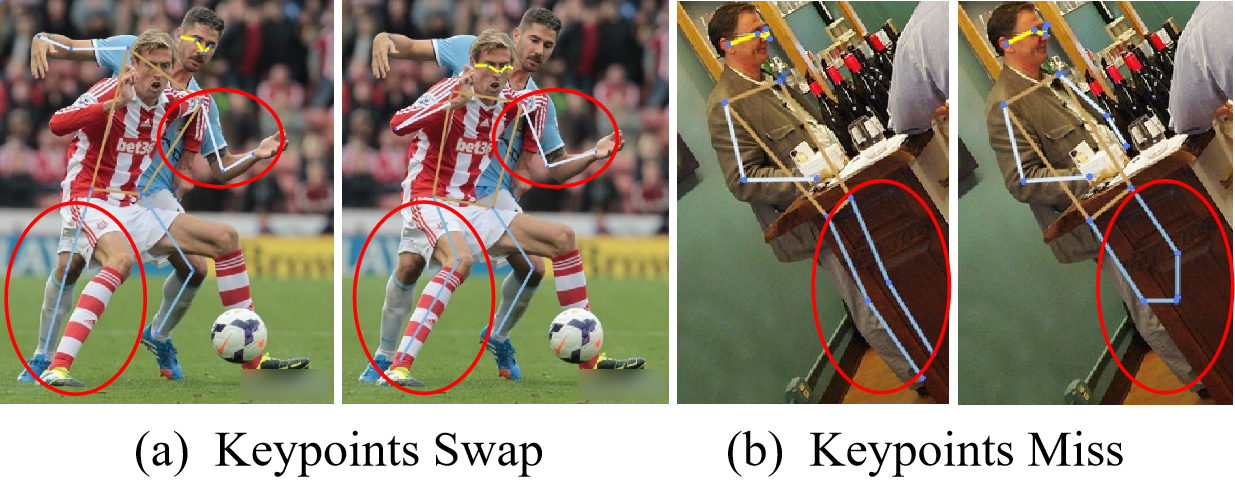}
  \caption{Challenges existed in occluded pose estimation. (a)Keypoints Swap: Due to the interference of the non-target person, some keypoints are predicted in the wrong positions. (b)Kypoints Miss: The large area of occlusion leads the contextual and structural information loss. Left is ground truth results and right is detection results.}
  \vspace{-0.3cm}
  \label{occlusion_problem}
\end{figure}
\setlength{\textfloatsep}{8pt}

Thanks to the powerful nonlinear mapping capabilities offered by neural networks, HPE has experienced notable advancements in recent years. However, existing methods such as PPNet\cite{zhao2021estimating}, TokenPose\cite{li2021tokenpose}, ViTPose\cite{xu2022vitpose} and PoseTrans\cite{jiang2022posetrans} all encounter challenges in addressing occlusion.
 

The primary difficulties of occluded human pose estimation mainly include the invisibility of occluded body parts and the strong interference caused by non-target keypoints. To be specific, there are three main factors: 1) Lack of occluded samples. Mainstream datasets of HPE lack occluded samples, which limits the ability of models to learn robust representations when facing occlusion. 2) Feature confusion. As shown in Fig.\ref{occlusion_problem}(a). Occlusion can lead to feature confusion because of the high similarity between the target and interfering persons, resulting in perplexity between the target and interfering keypoints. 3) Inference with uncompleted context information. As shown in Fig.\ref{occlusion_problem}(b). The large area occlusion leads to the loss of contextual and structural information, thus making the model unable to obtain enough contextual information from adjacent keypoints to infer the exact location, which leads to keypoints missing or abnormal postures.

\begin{figure*}[htb]
\vspace{-1em}
\centering
\includegraphics[width=0.8\textwidth]{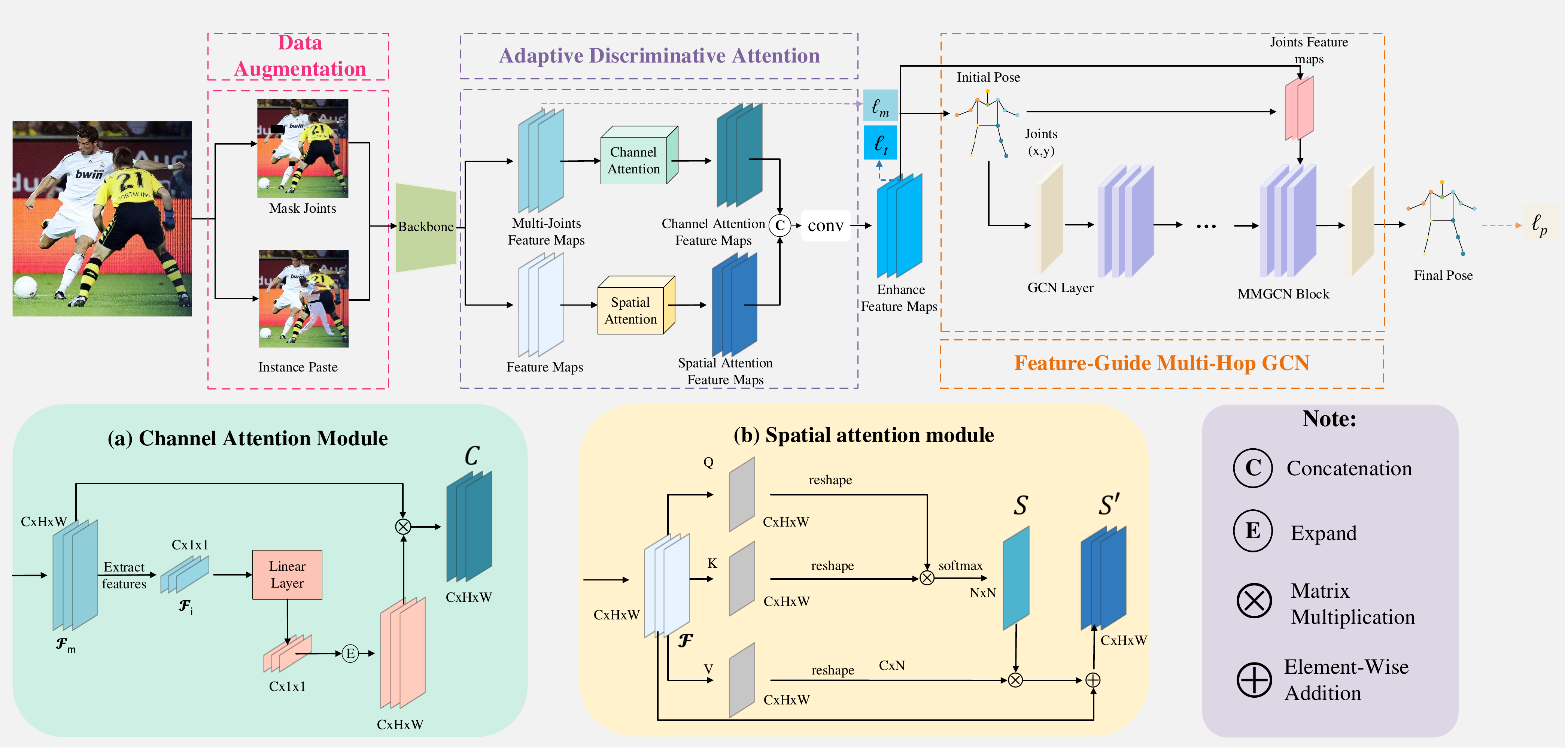}
\caption{The framework of our proposed method DAG. The input image undergoes data augmentation, then it is fed into the backbone for feature extraction. Subsequently, the features are input into the adaptive discriminative attention module for feature enhancement, the enhanced features are generated for initial pose generation. The initial pose is then sent to the feature-guided multi-hop GCN for pose refinement and correction, producing the final pose.} 
\label{pipeline}
\vspace{-1em} 
\end{figure*}

In response to the aforementioned challenges, several methods have been proposed. For instance, ASDA\cite{bin2020adversarial} and MSR-Net\cite{ke2018multi} try to generate occlusion samples, thereby mitigating the impact of limited occlusion samples in the dataset. STIP \cite{wang2021semantic} attempts to address occlusion by enhancing keypoints' semantic information. Besides, methods like OAS-Net\cite{zhou2020occlusion} focus on optimizing the processing of features to alleviate confusion between the target person and interfering individuals. Furthermore, to overcome the incomplete human body structure information caused by occlusion, approaches such as OPEC-Net\cite{qiu2020peeking} leverages the graph convolutional network (GCN) to infer occluded keypoints. However, these methods only address one aspect of the challenges, thus their performance is still not ideal when facing occlusion. 

In this paper, we present a comprehensive framework DAG (Data, Attention, Graph) to tackle the issue of occlusion in human pose estimation. First, we propose a novel data augmentation method called mask joints with instance paste. Unlike previous methods that focus on simulating either object occlusion or human occlusion alone, our method is compatible with both scenarios, allowing us to better simulate real-world occlusion situations. Second, in order to distinguish confused features caused by occlusion, an adaptive Discriminative Attention Module (ADAM) is introduced to differentiate target person and interfering individuals and enhance target features. Third, to compensate for incomplete body structure information, the Feature-Guided Multi-Hop GCN (FGMH-GCN) is introduced. FGMH-GCN can fully explore the prior knowledge of body structure and leverage useful information in the features map to compensate for the information loss in the initial pose estimation. Our method encompasses data augmentation, feature processing, and result refinement, providing a holistic approach to address occlusion-related difficulties.

\vspace{-0.5cm}
\section{Method}
\subsection{Overview}
As shown in Fig.\ref{pipeline}, our framework DAG is composed of three main components: Mask Joints with Instance Paste, Adaptive Discriminative Attention Module(ADAM), and Feature-Guided Multi-Hop GCN (FGMP-GCN). 

\vspace{-0.1cm}
\subsection{Mask Joints with Instance Paste}
We randomly mask partial keypoints to simulate object occlusion. On the other hand, person occlusion is simulated using instance paste, where occluding instances are inserted into the image.


\textbf{Mask Joints.} In scenarios where a person is occluded by an object, certain parts of the human body become invisible, so we use the joints mask method to simulate occlusion by objects, as shown in Fig.\ref{pipeline} . The process involves randomly selecting human and masking keypoints with rectangles of various sizes. This simulation enhances the robustness of the model by exposing it to occlusion challenges during training. 

\textbf{Instance Paste.} Inspired by ASDA\cite{bin2020adversarial} and in order to mitigate the adverse effects of person-to-person occlusion, we propose instance-paste to enhance the network's robustness. We begin by segmenting all human body images using the human parsing method\cite{ruan2019devil}. These segmented human bodies form our human body instances pool. Human instances are randomly selected from the instances pool with random rotation and scaling to create occlusion data.  Finally, the selected human instances are pasted into the image at random positions. This process generates diverse person patterns that mimic real-world scenarios. 

\subsection{Adaptive Discriminative Attention}
To address the challenge of feature confusion arising from occlusion, we introduce the  ADAM which comprises two components: channel attention and spatial attention.

\textbf{Channel Attention.} The channel attention leverages the central features of the human body to enhance the corresponding features, thereby facilitating the discrimination between different human instances at the channel level.

Given the multi-joints features $\mathcal{F}_m\in\mathbb{R}^{C\times H\times W}$ output from the backbone. H and W are the height and width of the features map, respectively. And the instance feature $\mathcal{F}_i\in\mathbb{R}^{C\times 1\times 1}$ is extracted from the center of the human body in $\mathcal{F}_m$. Then $F_i$ is fed into a linear layer to generate new features with the shape as $\mathcal{F}_i$. After that, an expansion operation is used to make the new features resemble $\mathcal{F}_m$. So the channel attention map $C \in\mathbb{R}^{C\times H\times W}$ can be defined as:
\begin{equation}
\setlength{\abovedisplayskip}{4pt} 
\setlength{\belowdisplayskip}{4pt}
C=\mathcal{F}_m\cdot \Psi(\mathbf{L}(\mathcal{F}_i)),
\end{equation}
$\mathbf{L}$ is a linear transformation, and $\Psi$ is the expansion operation.

\textbf{Spatial Attention.} Spatial attention is leveraged to suppress background features or object features that may occlude the human body.

Given the feature maps $\mathcal{F}\in\mathbb{R}^{C\times H\times W}$, we first use convolution layers to generate three new feature maps $Q, K, V$, where \{$Q, K, V$\} $\in\mathbb{R}^{C\times H\times W}$. Next, \{$Q, K$\} are reshaped to $\mathbb{R}^{C\times N}$, where $N=H\times W$ represents the number of pixels. Following this, the transpose of $K$ is multiplied by $Q$ to obtain the spatial attention map $S\in\mathbb{R}^{N\times N}$. After reshaping $V$ to $\mathbb{R}^{C\times N}$, a matrix multiplication is performed between $V$ and the transpose of $S$. The resulting matrix is then reshaped back to $\mathbb{R}^{C\times H\times W}$, yielding the final result $S'\in\mathbb{R}^{C\times H\times W}$:
\begin{equation}
\setlength{\abovedisplayskip}{4pt} 
\setlength{\belowdisplayskip}{4pt}
S'=\Phi {(\mathcal{F})_V} \cdot softmax((\Phi {(\mathcal{F})^T_K} \odot\Phi{(\mathcal{F})_Q}))^T.
\end{equation}
where $\Phi$ is a convolution operation. 

\subsection{Feature-Guided Multi-Hop Graph convolutional Network}
\textbf{GCN fomulation}  We construct a graph $G=(\mathcal{V},\mathcal{E})$ to formulate the human pose with N joints. Here, $\mathcal{V}$ represents each keypoint of the human body, and $\mathcal{E}$ represents the connections between two keypoints in the body. The collection of features of all nodes can be written as a matrix $M\in\mathbb{R}^{D\times N}$. The updated feature matrix can be written as:
\begin{equation}
\setlength{\abovedisplayskip}{4pt} 
\setlength{\belowdisplayskip}{4pt}
M' = \sigma(\sum_{k=1}^K{w_k}\cdot (W^{(0)}M+W^{(1)}M\hat{A}_k)).
\end{equation}
Where $w_k\in\mathbb{R}^{D\times N}$ is a learnable modulation matrix used to model the relationship between features at each hop distance, $W^{(0)}$ and $W^{(1)}$ are the weight matrices corresponding to the self and neighbor transformations, respectively. $\hat{A}$ is the symmetrically normalized version without self-connections. D is the dimension of features.

\textbf{Feature-Guided.} The MMGCN block \cite{lee2022multi} is employed to capture valuable information present in the features map that may have been lost in the initial pose. Next, the grid sampling is utilized to extract joint features from their respective feature map locations $(x, y)$ . Then the joint features are weighted into the original features map. 


\textbf{Multi-Hop.} Traditional graph convolutional networks (GCNs) often only consider information between adjacent nodes. Inspired by MMGCN \cite{lee2022multi}, we introduce the mechanism of multi-hop. The relationship in the adjacency matrix of each hop is low correlation except for the distance within k hops of the intermediate nodes. Therefore, it can provide a flexible modeling structure for learning the long-term relationships between human joints. FGMH-GCN not only makes up for the missing human structure information by using the prior knowledge of body structure but also suppresses the generation of abnormal poses.

\vspace{-0.1cm}
\subsection{Loss Function}
The loss function consists of three parts 1) multi-joints loss $\ell_m$; 2) target-joints loss $\ell_t$; 3) GCN-pose loss $\ell_p$. For the keypoint heatmap, we use mean square error (MSE) as our loss function. For the GCN pose loss, we use $L_1$ loss. So the overall loss function can be written as:
\begin{equation}
\setlength{\abovedisplayskip}{4pt} 
\setlength{\belowdisplayskip}{4pt}
\mathcal{L} = \ell_m + \ell_t + \lambda\ell_p.
\end{equation}
Where $\lambda$ is a hyperparameter for balancing the three losses.
\vspace{-0.2cm}

\begin{table*}[htbp]
\centering
\setlength{\tabcolsep}{10pt}
\caption{Comparison with Other methods on MSCOCO val, test-dev, RE, OCHuman, CrowdPose. The \underline{underlined} highlights the compared results, the results of our method are highlighted in \textbf{bold}.}
\vspace{-0.5cm}
\label{results}
\begin{tabular}{c|c|c|c|c|c|c|c}
\hline
\multirow{2}{*}{method} & \multirow{2}{*}{backbone} & \multirow{2}{*}{input size} &  val2017  & test-dev  & RE & OCHuman & CrowdPose\\
\cline{4-8} & & & $AP$ &$AP$  &$AP$  &$AP$  & $AP$   \\
\hline
SBL \cite{xiao2018simple} & ResNet-50 & 256x192  &70.4 & 70.0  & 67.1 & 55.8 & 68.4\\
OAS-Net \cite{zhou2020occlusion} & HRNet-W32 & 256x192 & 75.0 &- & -& -&-\\
ASDA \cite{bin2020adversarial} & HRNet-W32 & 256x192 & 75.2  &- & -& - & -\\
OPEC-Net \cite{qiu2020peeking} & HRNet-W32 & 256x192& - & 73.9  & - & - & -\\
HRNet \cite{sun2019deep} & HRNet-W32 & 256x192  &\underline{74.4}  & \underline{73.5}  & \underline{71.0} & 63.0 & \underline{71.7}\\
HRNet \cite{sun2019deep} & HRNet-W48 & 384x288  &\underline{76.3}  & \underline{75.5} & \underline{73.0} & 64.8 & \underline{73.9}\\
STIP \cite{wang2021semantic} & HRNet-W48 & 384x288 &76.8   & - & - &64.0 & -\\
PoseTrans \cite{jiang2022posetrans} & HRNet-W48 & 384x288 &76.8   &75.7  & - & - & -\\
SimCC \cite{li2022simcc} & HRNet-W32 & 256x192 &75.3  & 74.3  & 71.8 & \underline{62.3} & 66.7\\
SimCC \cite{li2022simcc} & HRNet-W48 & 384x288 &76.9 & 76.0  & 73.5 & \underline{66.2} & -\\
Poseur \cite{mao2022poseur} & ResNet-50 & 256x192 & \underline{74.2} & \underline{72.8}  & \underline{70.6} & \underline{58.0} & \underline{70.9}\\
\hline
\textbf{DAG} & HRNet-W32 & 256x192  &\textbf{75.4}\textcolor{red}{$\uparrow_{1.0}$}  & \textbf{74.3}\textcolor{red}{$\uparrow_{0.8}$} & \textbf{72.0}\textcolor{red}{$\uparrow_{1.0}$} &  \textbf{64.5}\textcolor{red}{$\uparrow_{2.2}$}& \textbf{72.7}\textcolor{red}{$\uparrow_{1.0}$} \\
\textbf{DAG} & HRNet-W48 & 384x288 &\textbf{77.0}\textcolor{red}{$\uparrow_{0.7}$} & \textbf{76.0}\textcolor{red}{$\uparrow_{0.5}$} & \textbf{73.7}\textcolor{red}{$\uparrow_{0.7}$}& \textbf{66.9}\textcolor{red}{$\uparrow_{0.7}$} & \textbf{74.2}\textcolor{red}{$\uparrow_{0.3}$} \\
\textbf{Poseur + DAG} & ResNet-50 & 256x192 &\textbf{74.5}\textcolor{red}{$\uparrow_{0.3}$} & \textbf{73.1}\textcolor{red}{$\uparrow_{0.3}$} & \textbf{71.1}\textcolor{red}{$\uparrow_{0.5}$}& \textbf{58.9}\textcolor{red}{$\uparrow_{0.9}$} &  \textbf{71.5}\textcolor{red}{$\uparrow_{0.6}$} \\
\hline
\end{tabular}
\vspace{-0.1cm}
\end{table*}
\vspace{-0.1cm}

\section{Experiments}
\vspace{-0.1cm}
\subsection{Datasets}
We evaluate our model on three datasets: MSCOCO-RE, 
CrowdPose \cite{li2019crowdpose} and OCHuman \cite{zhang2019pose2seg}.

\textbf{COCO-RE} the train2017 dataset contains 57K images with over 150K person instances. val2017 set and test-dev set contain 5K images and 20K images, respectively. COCO-RE is our re-labeling of the val2017 set, which adds the annotations of the occluded joints. The visible flag is denoted as $v=3:$ labeled and occluded.

\textbf{CrowdPose} \cite{li2019crowdpose} contains 20K images and 80K persons labeled with 14 keypoints. We use the trainval set for training and the test set for evaluation.

\textbf{OCHuman} \cite{zhang2019pose2seg} contains 4731 images and 8110 person instances. It consists of 2,500 images in the val set and 2,231 images in the test set.

\subsection{Implementation Details}
All experiments are conducted using PyTorch on two RTX 3090 GPUs. The HRNet is used as the baseline model and is initialized with weights pre-trained on the ImageNet classification task.

\textbf{Training.} We train all models using the HRNet framework, wherein the human bounding box is extended in a fixed aspect ratio of height:width=4:3, and the region is cropped from the original image which is then resized to a fixed size of 256x192 or 384x288. We follow the HRNet for other related settings.

\textbf{Testing.} We follow the top-down workflow for human pose estimation. In the case of the MSCOCO dataset, we use the detection results of previous works\cite{sun2019deep} to ensure a fair comparison. As for the CrowdPose dataset, we utilize ResNet101-FPN\cite{lin2017feature} as the human detector to detect individuals. The heatmap post-processing is the same as the HRNet.

\vspace{-0.35cm}
\subsection{Comparison with Other Methods}
Quantitative results of our method DAG and other methods on MSCOCO, OCHuman and CrowdPose are listed in Tab.\ref{results}. Compared to HRNet, DAG achieves 0.7\% and 0.5\% gains on val and test-dev set with HRNet-W48 and input size 384x288. The performance of traditional methods, such as SBL and HRNet, decreases when testing on MSCOCO-RE. In contrast, our DAG achieves a performance of 73.7 AP, which is an improvement of 0.7\% mAP compared to HRNet.

On OCHuman, We achieved 2.1\% mAP improvement compared to the baseline HRNet-W48 model with an input size of 384x288. Additionally, we achieve 2.2\% mAP improvement compared to the Simcc HRNet-W32 model with an input size of 256x192.

On CrowdPose, our proposed DAG has demonstrated a significant improvement of 1.0\% mAP compared to the baseline, indicating its effectiveness and generalizability across different scenarios.

Moreover, when combined with our DAG, the Poseur\cite{mao2022poseur} exhibited notable improvements across multiple datasets, especially in the occluded datasets. This provides strong evidence of DAG's effectiveness, which also shows DAG can be easily integrated into other frameworks, making it a versatile solution for improving human pose estimation performance in occlusion scenarios.
\vspace{0.5cm}

\begin{table}[htbp]
\centering
\setlength{\tabcolsep}{4pt}
\vspace{2pt}
\caption{Ablation study. Investigating the effects of the proposed module. DA means adding the proposed data augmentation module.
}
\label{ablation_study_coco}
\begin{tabular}{|c|c|c|c|c|c|}
\hline
\multicolumn{5}{|c|}{MSCOCO val2017 datsets} \\
\hline
HRNet-W32   & DA & ADAM & FGMH-GCN & $AP$  \\
\hline
$\checkmark$ &  &  &  & 74.4  \\
$\checkmark$ & $\checkmark$  &  &  & 74.7 ($0.3 \uparrow$) \\
$\checkmark$ &  & $\checkmark$  &  & 74.8 ($0.4 \uparrow$)\\
$\checkmark$ &  &  &$\checkmark$  & 75.1 ($0.7 \uparrow$)\\
$\checkmark$ &$\checkmark$  &$\checkmark$  &$\checkmark$  & 75.4 ($1.0 \uparrow$)\\
\hline
\end{tabular}
\end{table}

\vspace{-20pt}
\subsection{Ablation Study}
\noindent
\textbf{Ablation Studies on MSCOCO.} 
The results on MSCOCO are summarized in Tab.\ref{ablation_study_coco}. HRNet is used as the baseline with an input size of 256x192. Our proposed method improves
the mAP by 1\% compared to the baseline. Specifically, the data augmentation technique effectively simulates real-world occlusion scenarios and improves the mAP by 0.3\%. Moreover, ADAM distinguishes the target from the interference and further enhances the performance by 0.4\% mAP. Furthermore, FGMH-GCN improves the mAP by 0.7\%, effectively capturing the relationship between neighboring keypoints. The improvement induced by the proposed techniques clearly demonstrates the effectiveness of the proposed method.
%
\section{CONCLUSION}
\vspace{-10pt}

In this paper, we propose a comprehensive framework DAG for occluded pose estimation. In particular, we design a novel data augmentation method mask joints with instance paste to generate challenging occluded samples. Additionally, an Adaptive Discriminative Attention Module is introduced to distinguish the confusion features. Furthermore, we incorporate a Feature-Guided Multi-Hop GCN that leverages the human body structure to strengthen the body structure constraints during joint inference. Extensive experiments on three benchmarks demonstrate the effectiveness and generalizability of our method.

\section{ACKNOWLEDGMENTS}
This work was supported by the National Natural Science Fund of China under Grant No.62172222, 62072354, and the Postdoctoral Innovative Talent Support Program of China under Grant 2020M681609.

\bibliographystyle{IEEEbib}
\bibliography{refs}
\end{document}